\documentclass[11pt]{article}
\usepackage[parfill]{parskip}
\usepackage{graphicx}
\usepackage[top=1in, bottom=1in, left=1in, right=1in]{geometry}
\usepackage{amsmath}
\usepackage{hyperref}
\usepackage{fancyhdr}
\usepackage{titlesec}
\titlespacing{\section}{0pt}{\parskip}{-.5\parskip}
\titlespacing{\subsection}{0pt}{\parskip}{- .5\parskip}
\titlespacing{\subsubsection}{0pt}{\parskip}{- .5\parskip}

\date{\vspace{-5ex}}

\usepackage{authblk}
\author[]{Paul R. Cohen}
\affil[]{School of Computing and Information, University of Pittsburgh}

\usepackage{listings}
\usepackage{color}

\usepackage[subdued,defaultmathsizes]{mathastext}
\MTnonlettersobeymathxx     
\MTexplicitbracesobeymathxx 
\MTfamily {\ttdefault}      
\Mathastext [typewriter]    

\definecolor{dkgreen}{rgb}{0,0.6,0}
\definecolor{gray}{rgb}{0.5,0.5,0.5}
\definecolor{mauve}{rgb}{0.58,0,0.82}

\def\calF/{${\cal F}$}
\def\calR/{${\cal R}$}
\def\pram/{\textsc{pram}}
\def\adams/{\texttt{Adams}}
\def\berry/{\texttt{Berry}}
\def\home/{\texttt{Home}}

\def\g/{\texttt{g}}

\newenvironment{inl}[1]
    {$\mathtt{#1}$}

\newenvironment{grp}[1]
    {$\mathtt{g_{#1}}$}

\titlespacing{\align}{0pt}{\parskip}{- .95\parskip}

\title{Probabilistic Relational Agent-based Models (\textsc{pram})}

\begin{document}
\maketitle

\vspace{10pt}
\begin{abstract} 
\noindent \pram/ puts agent-based models on a sound probabilistic footing as a basis for integrating agent-based and probabilistic models.  It extends the themes of probabilistic relational models and lifted inference to incorporate dynamical models and simulation. It can also be much more efficient than agent-based simulation.  
\end{abstract}

\section{Introduction}

In agent-based models (ABMs, e.g.,~\cite{Kravari2015,Grefenstette2013}) agents probabilistically change state.  State can be represented as attribute values such as health status, monthly income, age, political orientation, location and so on. A population of agents has a {\em joint state} that is typically a joint distribution; for example, a population has a joint distribution over income levels and political beliefs.  ABMs are a popular method for exploring the {\em dynamics} of joint states, which can be hard to estimate when attribute values depend on each other, and populations are heterogeneous in the sense that not everyone has the same distribution of attribute values, and the principal mechanism for changing attribute values is interactions between agents. For example, suppose all agents have a flu status attribute that changes conditionally -- given other attributes such as age, income, and vaccination status -- when agents interact.  The dynamics of flu -- how it moves through heterogeneous populations -- can be difficult or impossible to solve, but ABMs can simulate the interactions of agents, and the flu status of these agents can be tracked over time.    

ABMs are no doubt engines of probabilistic inference, but it is difficult to say anything about the models that underlie the inference.  This paper presents \pram/ -- Probabilistic Relational Agent-based Models -- a new kind of ABM with design influences from compartmental models (e.g.,~\cite{Blackwood2018}) and probabilistic relational models (PRMs; e.g.,~\cite{Getoor2007}).  \pram/ seeks to clarify the probabilistic inference done by agent-based simulations as a first step toward integrating probabilistic and agent-based methods, enabling new capabilities such as automatic compilation of probabilistic models from simulation specifications, replacing or approximating expensive simulations with inexpensive probabilistic inference, and unifying ABMs with important methods such as causal inference.  

\section{An Example}

Consider the spread of influenza in a population of students at two schools, \adams/ and \berry/. To simplify the example, assume that flu spreads only at school.  Many students at \adams/ have parental care during the day, so when they get sick they tend to recover at home.  Most students at \berry/  lack parental care during the day, so sick students go to school.  Students may be susceptible, exposed or recovered.  

Several {\em groups} are entailed by this example, each of which has a {\em count}.  For example, at a point in time early in the flu season, the groups might look like those in Table~\ref{tab:simple}. Each school has 1000 students, but they are distributed differently: Compared with \adams/, there are more exposed students at \berry/ and fewer of them are at home.  At a later time, some exposed students will be recovered, some recovered students will again be susceptible, some home-bound students will be back at school, and in general the counts of groups will change.

\begin{table}[htb]
  \centering  
\texttt{
\begin{tabular}{lclr}
\hline
has\_school & flu\_status & has\_location & Count\\\hline
Adams & s & Adams & 697\\
Adams & s & Home & 3\\ 
Adams & e & Adams & 20\\
Adams & e & Home & 80\\
Adams & r & Adams & 180\\
Adams & r & Home & 20\\ \hline
Berry & s & Berry & 575\\
Berry & s & Home & 0\\
Berry & e & Berry & 330\\
Berry & e & Home & 10 \\
Berry & r & Berry & 115\\
Berry & r & Home & 0 \\
\hline
\end{tabular} 
}
\caption{Groups and counts at a specific time in a \pram/ simulation.}\label{tab:simple}
\end{table}

The central problem solved by \pram/ is to calculate how counts of groups change over time given the actions of {\em rules} that probabilistically change features and relations. For example, a rule might say: If group \g/ has daytime parental care and is exposed to flu, then change the location of \g/ to \g/'s home with probability $0.8$ and to \g/'s school with probability $0.2$. 

The connection between \pram/ models and probabilistic models is that counts are proportional to posterior probabilities conditioned on attributes such as \inl{has\_school} and \inl{flu\_status} and on the actions of rules that change attributes. \pram/ applies rules repeatedly to groups, creating novel groups and merging identical groups, thereby simulating the dynamics of groups' counts.   

\section{Elements of PRAM Models}

\pram/ entities have two kinds of attributes.  They have features, \calF/, which are unary predicates such as \inl{flu\_status} or \inl{income}; and they have relations, \calR/, between entities, such as \inl{has\_school}. Currently, \pram/ entities are {\em groups} and {\em sites}, and all forward relations relate one group to one site. Inverse relations relate one site to a set of groups.  Thus, if \inl{g_1.has\_school = Adams} and \inl{g_2.has\_school = Adams}, the inverse relation \inl{Adams.school\_of} returns \inl{\{g_1,g_2\}}.  Inverse relations are important for answering queries such as ``which groups attend \grp{1}'s school?"  Formally this would be \inl{g_1.has\_school.school\_of}, which would return \inl{\{g_1,g_2\}}.  By mapping over entities it is easy to answer queries such as ``what is the proportion of students at \grp{1}'s school that has been exposed to flu?"  In effect, \pram/ implements a simple relational database.

Besides entities, \pram/ models have rules that apply to groups.  All rules have mutually exclusive conditions, and each condition is associated with a probability distribution over mutually exclusive and exhaustive conjunctive actions. Thus, a rule will return exactly one distribution of conjunctive actions or nothing at all if no condition is true. For an illustration, look at the mutually exclusive clauses of  \inl{rule\_flu\_progression} in Figure~\ref{rules}, and particularly at the middle clause:  It tests whether the group's \inl{flu\_status == e} (exposed to flu) and it specifies a distribution over three conjunctive actions.  The first, which has probability $0.2$, is that the group recovers {\em and} becomes happy (i.e., change \inl{flu\_status} to {\tt r} {\em and} change \inl{mood} to {\tt happy}).  The remaining probability mass is divided between remaining exposed and becoming bored, with probability $0.5$, and remaining exposed and becoming annoyed, with probability $0.3$.

\begin{figure}
\begin{center}
\begin{lstlisting}
def rule_flu_progression (group):
    flu_status = group.get_feature('flu')
    location = objects_related_by(group,'has_location')
    infection_probability = location.proportion_located_here([('flu','e')])

    if flu_status == 's':
        return ((infection_probability,
                 ('change_feature','flu','e'),('change_feature','mood','annoyed')),
                ((1 - infection_probability),('change_feature','flu','s')))  
    elif flu_status == 'e':
        return ((.2, ('change_feature','flu','r'),('change_feature','mood','happy')),
                 (.5, ('change_feature','flu','e'),('change_feature','mood','bored')), 
                 (.3, ('change_feature','flu','e'),('change_feature','mood','annoyed')))   
    else flu_status == 'r':
        return ((.9, ('change_feature','flu','r')),
                 (.1, ('change_feature','flu','s')) 
                 
def rule_flu_location (group):
    ...        
    if flu_status == 'e' and income == 'l':
        return ((.1, ('change_relation','has_location',location,home)),
                 (.9, ('change_relation','has_location',location,location)))    
    elif flu_status == 'e' and income == 'm':
        return ((.6, ('change_relation','has_location',location,home)),
                 (.4, ('change_relation','has_location',location,location)))              
    else flu_status == 'r':
        return ((.8, ('change_relation','has_location',location,school)),
                 (.2, ('change_relation','has_location',location,location)))

\end{lstlisting}
\caption{Two \pram/ rules.  \inl{Rule\_flu\_progression} changes the \inl{flu\_status} and \inl{mood} features of a group. \inl{Rule\_flu\_location} changes a group's \inl{has\_location} relation.}
\label{rules}
\end{center}
\end{figure}

Next, consider the preamble of \inl{rule\_flu\_progression}, which queries the group's flu status, then finds the group's location, and then calls the method \inl{proportion\_located\_here} to calculate the proportion of flu cases at the location. (\inl{Proportion\_located\_here} sums the counts of groups at the location that have flu, then divides by the sum of the counts of all the groups at the location.)  In the rule's first clause, this proportion serves as a probability of infection. It is evaluated anew whenever the rule is applied to a group.  In this way, rules can test conditions that change over time. 
Finally, the third clause of the rule represents the  transition from \inl{flu\_status = r} back to \inl{flu\_status = s}, whereupon re-exposure becomes possible. 

In addition to changing groups' features, rules can also change relations such as {\tt has\_location}. The second rule in Figure~\ref{rules} says, if a group is exposed to flu and is {\em low}-income then change the group's location from its current \inl{location} to \inl{home} with probability $0.1$ and stay at \inl{location} with probability $0.9$.  If, however, the group is exposed and is {\em middle}-income, then it will go home with probability $0.6$ and stay put with probability $0.4$.  And if the group has recovered from flu, whatever its income level, then it will go back to school with probability $0.8$.
    
\section{The PRAM Engine:  Redistributing Group Counts}
\label{sec:engine}
    
The primary function of the \pram/ engine is to {\em redistribute} group counts among groups, as directed by rules, merging and creating groups as needed, in a probabilistically sound way. 
 
\pram/ groups are defined by their features and relations in the following sense:  Let \calF/ and \calR/ be features and relations of group {\tt g}, and let $n$ be the count of {\tt g}.  For groups \grp{i} and \grp{j}, if \calF/$_i =$ \calF/$_j$ and \calR/$_i =$ \calR/$_j$, then \pram/ will merge \grp{i} with \grp{j} and give the result a count of $n_i + n_j$.  Conversely, if a rule specifies a distribution of $k$ changes to \calF/$_i$ (or \calR/$_i$) that have probabilities $p_1,p_2,...,p_k$, then \pram/ will create $k$ new groups with the specified changes to \calF/$_i$ (or \calR/$_i$) and give them counts equal to $(p_1 \cdot n_i), (p_2 \cdot n_i), ..., (p_k \cdot n_i)$. 

\paragraph{Redistribution Step 1: Potential Groups} To illustrate the details of how \pram/ redistributes counts, suppose in its initial conditions a \pram/ model contains just two {\em extant} groups:

\begin{center}
\texttt{
\begin{tabular}{lcccr}
name & flu & mood & location & count \\ \hline
\grp{1} & s & happy & adams & 900 \\
\grp{2} & e & annoyed & adams & 100 
\end{tabular}
}\end{center}
\vspace{.075in}

When {\tt rule\_flu\_progression} is applied to \grp{1} it calculates the \inl{infection\_probability} at \inl{Adams} to be $100 / (100 + 900) = .1$.  \grp{1} triggers the first clause in the rule because \grp{1}'s \inl{flu\_status == s}. So the rule specifies that the \inl{flu\_status} of \grp{1} changes to {\tt e} with probability $0.1$ and changes to {\tt s} with probability $0.9$.  \pram/ then creates two {\em potential groups}:

\begin{center}
\texttt{
\begin{tabular}{lcccr}
name & flu & mood & location & count \\ \hline
\grp{1\_1} & e & annoyed & adams & 90 \\
\grp{1\_2} & s & happy & adams & 810 
\end{tabular}
}\end{center}
\vspace{.075in}

These potential groups specify a {\em redistribution} of $n_1$, the count of \grp{1}. We will see how \pram/ processes redistributions, shortly. 

Of the two rules described earlier, {\tt rule\_flu\_location} does not apply to \grp{1}, but both apply to group \grp{2}. When multiple rules apply to a  group, \pram/ creates the cartesian product of their distributions of actions and multiplies the associated probabilities accordingly, thereby enforcing the principle that rules' effects are independent. (If one wants dependent effects they should be specified {\em within} rules.) To illustrate, {\tt rule\_flu\_progression} specifies a distribution of three  actions for groups like \grp{2} that have \inl{flu\_status=e}, with associated probabilities $0.2,0.5,0.3$; while {\tt rule\_flu\_location} specifies two locations for groups that have \inl{flu\_status=e} and \inl{flu\_status=m}, with probabilities $0.6$ and $0.4$.  Thus, for \grp{2}, there are six joint actions of these two rules, thus six potential groups: 

\begin{center}
\texttt{
\begin{tabular}{lcccl}
name & flu & mood & location & count \\ \hline
\grp{2\_1} & r & happy & home & 100 $\cdot$ 0.2 $\cdot$ 0.6 = 12.0 \\
\grp{2\_2} & r & happy & adams & 100 $\cdot$ 0.2 $\cdot$ 0.4 = 8.0 \\
\grp{2\_3} & e & bored & home & 100 $\cdot$ 0.5 $\cdot$ 0.6 = 30.0 \\
\grp{2\_4} & e & bored & adams & 100 $\cdot$ 0.5 $\cdot$ 0.4 = 20.0 \\
\grp{2\_5} & e & annoyed & home & 100 $\cdot$ 0.3 $\cdot$ 0.6 = 18.0 \\
\grp{2\_6} & e & annoyed & adams & 100 $\cdot$ 0.3 $\cdot$ 0.4 = 12.0 \\
\end{tabular}
}\end{center}
\vspace{.075in}

These groups redistribute the count of \grp{2} (which is 100) by multiplying it by the product of probabilities associated with each action. 

\paragraph{Redistribution Step 2: The Redistribution Method} \pram/  applies all rules to all groups, collecting potential groups as it goes along.  Only then does it redistribute counts, as follows:  
\begin{enumerate}
\item Extant groups that spawn potential groups have their counts set to zero;
\item Potential groups that match extant groups (i.e., have identical \calF/s and \calR/s) contribute their counts to the extant groups and are discarded;
\item Potential groups that don't match extant groups become extant groups with their given counts.
\end{enumerate} 

So: Extant groups \grp{1} and \grp{2} have their counts set to zero.  Potential group \grp{1\_2} has the same features and relations as \grp{1} so it contributes its count, 810, to \grp{1} and is discarded.  Likewise, potential group \grp{1\_1} matches \grp{2} so it contributes 90 to \grp{2} and is discarded. Potential group \grp{2\_6} also matches \grp{2}, so it contributes 12 to \grp{2} and is discarded, bringing \grp{2}'s total to 102. Potential groups \grp{2\_1}, \grp{2\_2}, \grp{2\_3}, \grp{2\_4}, and \grp{2\_5} do not match any extant group, so they become extant groups.  The final redistribution of extant groups \grp{1} and \grp{2} is:

\begin{center}
\texttt{
\begin{tabular}{lcccr}
name & flu & mood & location & count \\ \hline
\grp{1} & s & happy & adams & 810.0 \\
\grp{2} & e & annoyed & adams & 102.0 \\
\grp{2\_1} & r & happy & home & 12.0 \\
\grp{2\_2} & r & happy & adams & 8.0 \\
\grp{2\_3} & e & bored & home & 30.0 \\
\grp{2\_4} & e & bored & adams & 20.0 \\
\grp{2\_5} & e & annoyed & home & 18.0 \\
\end{tabular}
}\end{center}

\vspace{.075in}


\paragraph{Redistribution Step 3: Iterate}  \pram/ is designed to explore the dynamics of group counts, so it generally will run iteratively.  At the end of each iteration, all non-discarded groups are marked as extant and the preceding steps are repeated: All rules are applied to all extant groups, all potential groups are collected, potential groups that match extant groups are merged with them, and new extant groups are created.  A second iteration produces one such new group when the third clause of {\tt rule\_flu\_progression} is applied to \grp{2\_1}:

\begin{center}
\texttt{
\begin{tabular}{lcccr}
name & flu & mood & location & count \\ \hline
\grp{2\_1\_1} & s & happy & home & 0.24 \\
\end{tabular}
}\end{center}

\vspace{.075in}

The reader is invited to calculate the full redistribution resulting from a second iteration (it is surprisingly difficult to do by hand).\footnote{The second iteration produces $n_1 = 706.632,~~n_2 = 119.768, ~~n_{2\_1} = 26.4, ~~n_{2\_1\_1} = 0.24, ~~n_{2\_2} = 25.6, ~~n_{2\_3} = 60.6, ~~n_{2\_4} = 24.4, ~~n_{2\_5} = 36.36$.}

\section{Exploring Population Dynamics with {\sc pram}} 

Extending an earlier example, suppose the schools \adams/ and \berry/ each enroll 1000 students, of whom  900 are susceptible and 100 are exposed, evenly divided between males and females.  All \adams/ students are middle-income and all \berry/ students are low-income. No students are pregnant, but we add a rule that creates pregnancies in groups of females with probability $0.01$.  The initial eight extant groups are:

\begin{center}
\texttt{
\begin{tabular}{lccccccr}
name & flu & sex & income & pregnant & mood & location & count \\ \hline
g1 & s & f & m & no & happy & adams & 450 \\
g2 & e & f & m & no & annoyed & adams & 50 \\
g3 & s & m & m & no & happy & adams & 450 \\
g4 & e & m & m & no & annoyed & adams & 50 \\
g5 & s & f & l & no & happy & berry & 450 \\
g6 & e & f & l & no & annoyed & berry & 50 \\
g7 & s & m & l & no & happy & berry & 450 \\
g8 & e & m & l & no & annoyed & berry & 50 
\end{tabular}
}\end{center}
\vspace{.075in}

\begin{figure}[hbt]
\begin{center}
\includegraphics[width=6in]{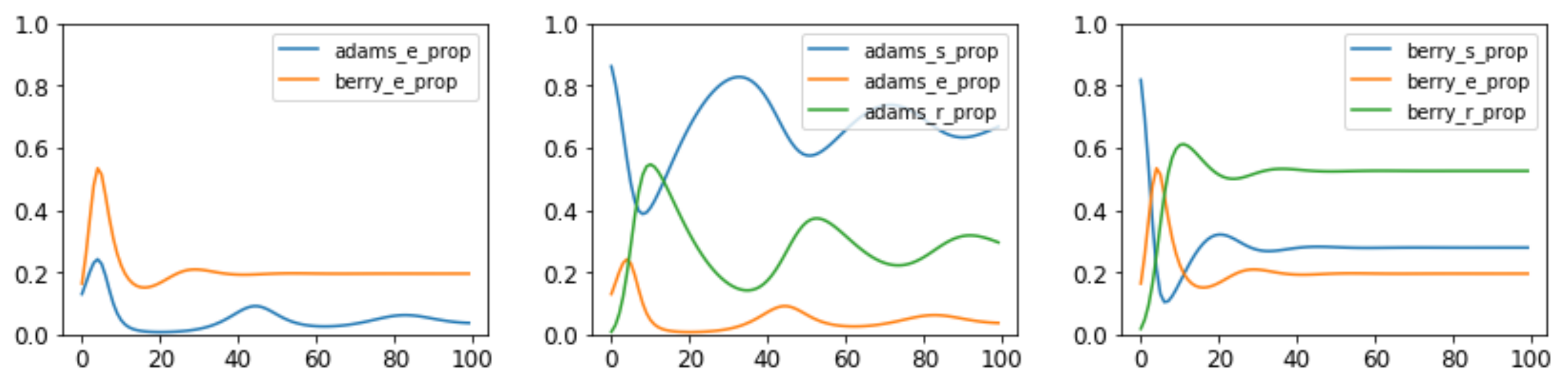}
\caption{The left panel shows the proportion of students exposed to flu at \adams/ and \berry/.  The middle and right panels show the proportions of susceptible, exposed and recovered students at each school. The simulation ran for 100 iterations.}
\label{fig:dynamics_e}
\end{center}
\end{figure}

The dynamics of \inl{flu\_status = e} at the two schools is presented in Figure~\ref{fig:dynamics_e}. The leftmost panel shows the proportion of students exposed to flu at each school.  \berry/ experiences a strong epidemic, with more than half the students exposed, whereas \adams/ has a more attenuated epidemic because its students are middle-income and can stay home when they are exposed, thereby reducing the infection probability at the school. {\tt Adams'} endemic level of flu is close to zero whereas \berry/'s endemic level is around 20\%.  However, resurgent flu caused by recovered cases becoming susceptible again is more noticeable at \adams/ (around iteration 45).  The only difference between \adams/ and \berry/ is that 60\% of \adams/ students stay home when they get flu, whereas 10\% of \berry/ students do, but this difference has large and persistent consequences. 

\section{Discussion}

\pram/ code is available on github ~\cite{Loboda2019}.  It has run on much larger problems, including a simulation of daily activities in Allegheny County that involved more than 200,000 groups.  \pram/ runtimes are proportional to $\nu$ the number of groups, not the group counts, so \pram/ can be much more efficient than agent-based simulations (ABS).  Indeed, when group counts become one, \pram/ {\em is} an ABS, but in applications where agents or groups are {\em functionally identical} \pram/ is more efficient than ABS. (Two entities $i$ and $j$ are functionally identical if \calF/$_i =$ \calF/$_j$ and \calR/$_i =$ \calR/$_j$ after removing all features from \calF/$_i$ and \calF/$_j$ and all relations from \calR/$_i$ and \calR/$_j$ that are not mentioned in any rule.) 

Because $\nu$ depends on the numbers of features and relations, and the number of discrete values each can have, \pram/ could generate enormous numbers of groups.  In practice, the growth of $\nu$ is controlled by the number of groups in the initial population and the actions of rules.  Typically, $\nu$ grows very quickly to a constant, after which \pram/ merely redistributes counts between these groups.  In the preceding example, the initial $\nu = 8$ groups grew to $\nu = 44$ on the first iteration and $\nu = 52$ on the second, after which no new groups were added.   

This dependence between $\nu$ and the actions of rules suggests a simple idea for {\em compiling} populations given rules: Any feature or relation that is not mentioned in a rule need not be in groups' \calF/ or \calR/.  Said differently, the only attributes that need to be in groups' definitions are those that condition the actions of rules. Currently we are building a compiler for \pram/ that automatically creates an initial set of groups from two sources: A database that provides \calF/ and \calR/ for individuals and a set of rules.  The compiler eliminates from \calF/ and \calR/ those attributes that aren't queried or changed by rules, thereby collapsing a population of individuals into groups with known counts.  

Attributes with continuous values obviously can result in essentially infinite numbers of groups. 
(Imagine one group with a single real-valued feature and one rule that adds a standard normal variate to it. Such a \pram/ model would double the number of groups on each iteration without limit.) 
Rather than ban real-valued attributes from \pram/ we are working on a method by which groups have distributions of such attributes and rules change the parameters of these distributions. We are developing efficient methods by which \pram/ generates new potential groups and tests whether they match extant groups.  

For all this talk of efficiency, the primary advantage of \pram/ over ABS is that \pram/ models are guaranteed to handle probabilities properly.  The steps described in Section~\ref{sec:engine} ensure that group counts are consistent with the probability distributions in rules and are not influenced by the order in which rules are applied to groups, or the order in which rules' conditions are evaluated. These guarantees are the first step toward a seamless unification of databases with probabilistic and \pram/ models.  The next steps, which we have already taken on a very small scale, are automatic compilation of probabilistic models given \pram/ models, and automatic compilation of \pram/ rules given probabilistic models. Probabilistic relational models, which inspired \pram/, integrate databases with lifted inference in Bayesian models; \pram/ adds simulation to this productive mashup, enabling models of dynamics.  

\section{Acknowledgments}
This work is funded by the DARPA program ``Automating Scientific Knowledge Extraction (ASKE)'' under Agreement HR00111990012 from the Army Research Office.


\begin{thebibliography}{9}
\bibitem{Blackwood2018} 
Julie C. Blackwood and Lauren M. Childs. An introduction to compartmental modeling for the budding infectious disease modeler. Letters in Biomathematics, 5(1), pp.195-221. 2018. doi:10.1080/23737867.2018.1509026

\bibitem{Getoor2007}
Lise Getoor, Ben Taskar (Eds.) Introduction to Statistical Relational Learning.  2007. MIT Press

\bibitem{Grefenstette2013}
Grefenstette JJ, Brown ST, Rosenfeld R, Depasse J, Stone NT, Cooley PC, Wheaton WD, Fyshe A, Galloway DD, Sriram A, Guclu H, Abraham T, Burke DS. FRED (A Framework for Reconstructing Epidemic Dynamics): An open-source software system for modeling infectious diseases and control strategies using census-based populations. BMC Public Health, 2013 Oct;13(1), 940. doi: 10.1186/1471-2458-13-940.

\bibitem{Kravari2015}
Kalliopi Kravari and Nick Bassiliades. A Survey of Agent Platforms.  2015.

\bibitem{Loboda2019} 
The version of \pram/ reported here was developed by the author.  A better engineered version has been developed by Tomek Loboda: \url{https://github.com/momacs/pram/} with documentation at \url{https://github.com/momacs/pram/blob/master/docs/Milestone-3-Report.pdf}
\end{thebibliography}
\end{document}